\documentclass[runningheads]{llncs}
\usepackage{graphicx}
\usepackage{makecell}
\usepackage{booktabs}
\usepackage{times}
\usepackage{epsfig}
\usepackage{amsmath}
\usepackage{amssymb}
\usepackage{multirow}
\begin{document}
\title{Spherical Transformer: Adapting Spherical Signal to Convolutional Networks}
%
%
 \author{
Yuqi Liu\inst{1} \and
 Yin Wang\inst{1} \and
 Haikuan Du\inst{2}\and
 Shen Cai\inst{*,2}
 }
 \institute{School of Electronics and Information Engineering, Tongji University, Shanghai, China\\
 \email{rhodaliu17@163.com, yinw@tongji.edu.cn} \and
 Visual and Geometric Perception Lab, Donghua University, Shanghai, China\\
 \email{\{Hanson\_du,hammer\_cai\}@163.com}}
%
\maketitle              
\renewcommand{\thefootnote}{}
\footnote{This work is supported by the Foundation of Key Laboratory of Artificial Intelligence, Ministry of Education, P.R. China (AI2020003).}

\vspace{-8mm}
\begin{abstract}
Convolutional neural networks (CNNs) have been widely used in various vision tasks, e.g. image classification, semantic segmentation, etc. Unfortunately, standard 2D CNNs are not well suited for spherical signals such as panorama images or spherical projections, as the sphere is an unstructured grid. In this paper, we present Spherical Transformer which can transform spherical signals into vectors that can be directly processed by standard CNNs such that many well-designed CNNs architectures can be reused across tasks and datasets by pretraining. To this end, the proposed method first uses local structured sampling methods such as HEALPix to construct a transformer grid by using the information of spherical points and its adjacent points, and then transforms the spherical signals to the vectors through the grid. By building the Spherical Transformer module, we can use multiple CNN architectures directly. We evaluate our approach on the tasks of spherical MNIST recognition, 3D object classification and omnidirectional image semantic segmentation. For 3D object classification, we further propose a rendering-based projection method to improve the performance and a rotational-equivariant model to improve the anti-rotation ability. Experimental results on three tasks show that our approach achieves superior performance over state-of-the-art methods.
\vspace{-2mm}
\keywords{Spherical transformer \and 3D object classification \and Omnidirectional image semantic segmentation.}
\end{abstract}
%
%
%

\vspace{-8mm}
\section{Introduction}
Recently, with the increasing availability and popularity of omnidirectional sensors, a wide range of learning problems in computer vision and related areas require processing signals in spherical domain. For instance, for omnidirectional images and 3D models, standard CNNs working with structured data are not perfectly applicable to them. This limitation is especially pronounced for 3D models, which provides accurate encoding of 3D objects. In recent years, various methods have been proposed to deal with 3D objects for different data representations, such as voxelization, point cloud, multi-view images and spherical images. 
Recent works~\cite{s.2018spherical,10.1007/978-3-030-01261-8_4} propose different network architectures that achieve rotation invariance and direct operation in spherical domain. However, the sampling method in these networks is preliminary, which only maps spherical signals to local-distorted planar domains. This will result in undesirable distortions and loss of information. 
The work~\cite{jiang2018spherical} presents a new convolutional kernel to apply CNNs on unstructured grids, and overcomes the above shortcomings. However, it does not contain the standard convolutional operation which shows strong capabilities in 2D image classification and semantic segmentation tasks. Seeing the rich asset of existing CNNs modules 
for traditional image data, we are motivated to devise a mechanism to bridge the mainstream standard CNNs to spherical signals. One particular challenge is how to express spherical signal in a structured manner. 

Classic deep neural networks take structured vectors as input. How to encode the input vectors is crucial to the performance. Spatial Transformer Networks (STN)~\cite{jaderberg2015spatial} are commonly used to learn spatial transformations on the input image to enhance the geometric invariance of the model. Inspired by STN, we present a spherical transformer method to transform the spherical signals to structured vectors that can be directly processed by standard CNNs. Therefore, it is able to deal with different tasks of spherical signals. In this paper, we focus on three typical applications: spherical MNIST recognition, 3D model classification and omnidirectional images semantic segmentation. In detail, we employ the popular VGG-11~\cite{simonyan2014very} and U-Net~\cite{ronneberger2015u} for 3D model classification and semantic segmentation separately. For MNIST recognition on sphere, we use a simple CNN with 5 layers to accomplish this task. 

In summary, this paper makes the following contributions:
\vspace{-2mm}
\begin{itemize}
\item We devise a spherical transformer module which can transform spherical signals into vectors friendly to standard CNNs. 
We can easily reuse the VGG-11~\cite{simonyan2014very} and U-Net~\cite{ronneberger2015u} architecture for spherical data. Moreover, our seamless use of standard CNNs enables the benefit of pretraining (transfer learning) across different datasets.
\item We propose a novel rendering-based projection method which can project the local shading of the internal object points to the spherical image. 
 This rendering method can enrich the spherical features and improve the classification performance of 3D objects.
\item Our spherical CNNs achieve superior performance on spherical MNIST recognition, 3D object classification and spherical image semantic segmentation, compared to state-of-the-art networks~\cite{s.2018spherical,jiang2018spherical}
\end{itemize}

\vspace{-6mm}
\section{Related Work}
\vspace{-2mm}
\label{sec:related}
\paragraph{3D-based methods.}
3D-based methods are mainly designated for voxel grids or point clouds as input data format. Like 2D images, voxels can be directly processed by using 3D convolution operations. However, due to memory and computational restrictions, the voxel-based methods are often limited to small voxel resolution. 
This will result in the loss of many details. Although this issue has been recently alleviated by Octree-based representations such as OctNet~\cite{Riegler2017OctNet}, the cost of 3D CNNs is still distinctly higher than 2D neural networks of the same resolution. 

Alternative approaches have been recently explored to handle 3D points, such as PointNet~\cite{qi2017pointnet}.
It proposes a novel network architecture that operates directly on point clouds. Moreover, PointNet uses the global max pooling to solve the disorder of the point cloud input. However, PointNet cannot capture the local structure, which limits its ability to identify fine scenes and generalize complex scenes. PointNet++~\cite{qi2017pointnet++} and DGCNN~\cite{wang2018dynamic} alleviate this problem and achieve improved performance. SubdivNet\cite{SubdivNet} also adopts 2D CNN to process 3D Mesh by exploiting an analogy between mesh faces and 2D images, which is similar to our idea. Despite this, our method is proposed earlier, although it uses specific Spherical data.
\newcommand{\tabincell}[2]{\begin{tabular}{@{}#1@{}}#2\end{tabular}}

\begin{table*}[tb!]
\caption{Features of 3D object recognition networks.\centering}
  \label{method_table}
  \centering
	\resizebox{0.7\textwidth}{!}
	{
    \begin{tabular}{lcc}
    \toprule
    Method     & Input     & Convolution feature \\
    \midrule[1pt]
    Voxnet \cite{maturana2015voxnet}       & \multirow{2}{*}{voxels}  & \multirow{2}{*}{3D CNNs}\\
    OctNet~\cite{Riegler2017OctNet}    &  & \\
    \midrule[1pt]
    PointNet~\cite{qi2017pointnet} & \multirow{4}{*}{points}  & \multirow{4}{*}{\tabincell{c}{MLP implemented\\ by $1\times1$ conv kernel} }\\
    PointNet++~\cite{qi2017pointnet++}   &  & \\
    DGCNN \cite{wang2018dynamic}    &   & \\
    SFCNN \cite{rao2019spherical}    &   & \\
    \midrule[1pt]
    MVCNN~\cite{su15mvcnn} & \multirow{2}{*}{images}  & \multirow{2}{*}{2D CNNs}\\
    MVCNN-NEW~\cite{Su2018ADL}   &   & \\
    \midrule[1pt]
    S2CNN \cite{s.2018spherical}  & \multirow{3}{*}{spherical}  & \multirow{2}{*}{\tabincell{c}{FFT of spherical \\ non-uniform points}}\\
    SphericalCNN~\cite{10.1007/978-3-030-01261-8_4}       &   &  \\
    \cmidrule{3-3}
    UGSCNN~\cite{jiang2018spherical}            &   & \tabincell{c}{parameterized differential \\operators to replace 2D conv}   \\
    \midrule
    DeepSphere\cite{perraudin2019deepsphere}            & spherical   &  GNNs implemented by $1\times1$ conv kernel   \\
    \midrule
    STM (Ours)             & spherical   &  2D CNNs   \\
    \bottomrule
  \end{tabular}}
  \vspace{-4mm}
\end{table*}

\vspace{-3mm}
\paragraph{Planar-image-based methods.}
One advantage of 3D based technology is that 3D ob\-jects can be accurately described by the corresponding 3D representations. However, the network architecture applying to 3D representations is (arguably) still in its open stage. As standard CNNs have achieved great success in the field of planar images, there exist methods using 3D object's multi-view images to classify the object. The two prevailing methods are MVCNN~\cite{su15mvcnn} and MVCNN-new~\cite{Su2018ADL}. They produce 12-view images of a 3D model and classify the rendered images through standard convolutional neural network architectures such as VGG~\cite{simonyan2014very} and ResNet \cite{He2016DeepRL} which can be pretrained on ImageNet dataset. 
Although image-based technology can take advantage of the pre-trained parameters of classification networks and achieve superb results, object projection inevitably leads to information loss in theory.

For semantic segmentation, there have been recently proposed many methods, such as U-Net~\cite{ronneberger2015u} and deeplab~\cite{Chen2018DeepLabSI} etc. Nevertheless, these methods are all devised for 2D images. Converting a spherical panorama directly into a equirectangular image leads to severe distortion, especially near the pole, which makes these methods less effective. 
\vspace{-3mm}

\paragraph{Spherical-image-based methods.}
Several other works e.g. spherical CNNs~\cite{s.2018spherical} seek to combine 3D-based techniques and image-based techniques. In the work~\cite{2017arXiv170201105A,chang2017matterport3d}, the authors provide benchmarks for semantic segmentation of 360 panorama images. 

For 3D object classification, several classification networks based on spherical depth projection are proposed in recent years.
\cite{s.2018spherical} proposes spherical convolutions that are rotational-equivariant. And spherical harmonic basis is used in \cite{10.1007/978-3-030-01261-8_4} to obtain similar results. The above methods have achieved good results on 3D object classification through depth-based spherical projection. However, they often have restrictive applicability and can only be used for a specific task, mainly due to the inflexibility of the adopted sampling method. More recently, convolution is achieved on unstructured grids by replacing the standard convolution with a differential operation~\cite{jiang2018spherical}. While this kind of pseudo convolution lacks the ability to reuse standard CNNs. SFCNN \cite{rao2019spherical} proposes a spherical fractal CNNs for point cloud analysis. It first projects the point cloud with extracting features onto the corresponding spherical points, and then uses the graph convolution for classification and object segmentation. 

For equirectangular image segmentation, \cite{su2017learning} proposes spherical convolution by changing the convolution kernel through equirectangular projection. SphereNet~\cite{coors2018spherenet} introduces a framework for learning spherical image representations by encoding distortion invariance into convolutional filters. But this method only supports $3\times3$ convolution kernel and max pool, and it needs bilinear interpolation to index the adjacent points of spherical points while we are directly looking for them. 

In recent studies, DeepSphere~\cite{perraudin2019deepsphere} proposes a spherical CNNs by using graph neural networks for cosmological data analysis. Table~\ref{method_table} briefly summaries the convolution characteristics of common 3D object recognition networks.

\vspace{-3mm}
\section{The Proposed Approach}
\label{sec:method}
\vspace{-2mm}
\subsection{Spherical sampling}
\label{sec31}
\begin{figure*}[tb!]
  \centering
  \includegraphics[width=\textwidth]{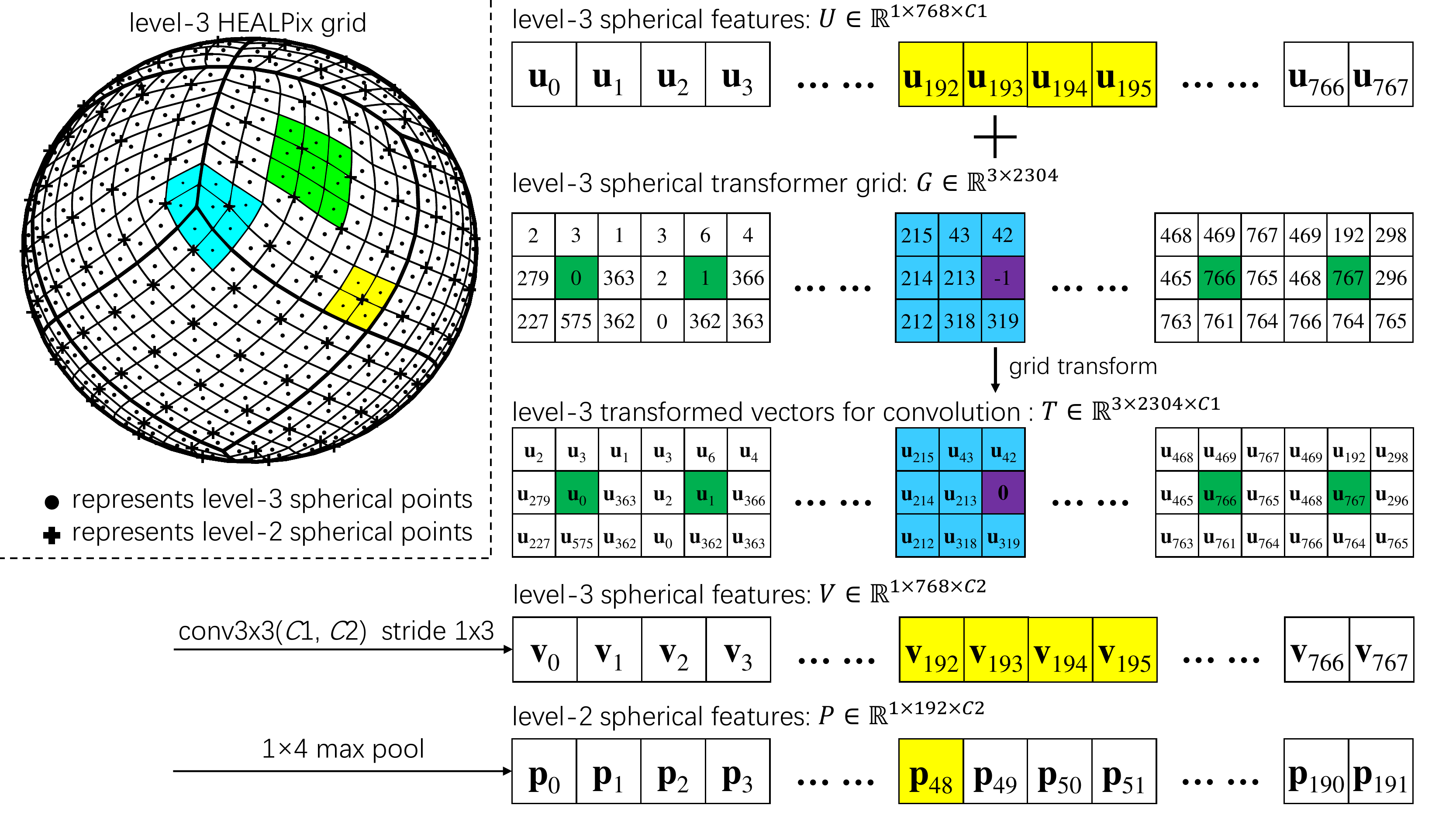}
  \vspace{-6mm}
  \caption{Sketch for how the spherical transformer works. The left side shows the level-\emph{3} HEALPix grid; the right and bottom show how we use the transformer grids to process spherical signals. For example, the four yellow dots in level-3 grid with index 192-195 will be transformed to the yellow cross with index 48 in level-2 grid after $1\times4$ max pool; For the green dot with eight adjacent points, such as index 0, 1, 766 and 767, the $3\times3$ spherical transformer grid can be constructed directly; While for the blue dot with seven adjacent points, such as index 213, the missing purple element is set to -1 in transformer grid and set to \textbf{0} in transformed vectors.}
  \label{fig:module}
  \vspace{-6mm}
\end{figure*}

The major concern for processing spherical signals is distortion introduced by projecting signals on other formats, such as projecting panorama image to planar image. Thus the best way to process spherical signals is finding a spherical uniform sampling method. We find HEALPix spherical grid~\cite{Gorski:2004by} (as shown in Fig.~\ref{fig:module}) and icosahedral spherical grid ~\cite{1985SJNA...22.1107B} methods are able to achieve this goal. Here we adopt HEALPix sampling in the proposed method, mainly because most spherical points on HEALPix grid have 8 neighbor points. It can form $3\times3$ transformer grid and then convert to vectors to be processed by $3\times3$ convolution kernels. But for icosahedral grid, all spherical points has 6 neighbors, which means they can only be processed by $1\times7$ convolution kernels. HEALPix grid starts with 12 points as the level-\emph{0} resolution. Each progressive level resolution after is 4 times the previous number of points. Hence, the number of sampling points on the sphere $n_{p}^{l}$ with level-\emph{l} is:
\begin{equation}
 n_{p}^{l} = 12\cdot4^\emph{l} \label{equ:1}
\end{equation}
It can be clearly discovered that the resolution changing process perfectly matches the $2\times2$ pooling layer. 

\vspace{-3mm}
\subsection{Spherical transformer module}
\label{sec32}
Sphere is local planar and a few of spherical grids in local region are structured, such as HEALPix spherical grid~\cite{Gorski:2004by} as adopted in the paper. Thus the next key step is to conduct a transformer grid by using the location information of spherical points and their neighbors. Specifically, our proposed spherical transformer module (STM) includes two layers: spherical convolution layer and spherical pooling layer, as will be described in detail. 
\vspace{-4mm}
\paragraph{Spherical convolution layer.}
Standard convolution layer needs structured grid such as image data. As shown in Fig.~\ref{fig:module}, the local points are structured in HEALPix spherical grid. It can be seen that most of the spherical points have eight adjacent points, one of which is depicted with green in Fig.~\ref{fig:module}. For level-\emph{l} (\emph{l} $\geq$1), there are always 24 points with the number of the adjacent points being 7. For specific examples, see the blue part of Fig.~\ref{fig:module}. For every point on HEALPix grid, the spherical convolution operation includes the following steps, as shown in Fig.~\ref{fig:module} for level-\emph{3} HEALPix grid:

\textbf{1) Determine the point index and extract the corresponding feature vectors.} Here we adopt the nested scheme which arranges the point index hierarchically (see~\cite{Gorski:2004by} for detail). 768 points in level-\emph{3} HEALPix grid are arranged from small to large as the first row of Fig.~\ref{fig:module} whose element is the feature vector of the current index point.

\textbf{2) Build the spherical transformer grid of convolution beforehand.} As the HEA\-LPix grid point index are preassigned, the indexes of the adjacent points for every point is known. Thus we can generate the $3\times3$ transformer grid for every point. For some points with only 7 adjacent points, the grid value of the missing position is set to be -1. The second row of Fig.~\ref{fig:module} shows some examples.

\textbf{3) Obtain the transformed vectors for convolution.} This transform operation is very similar to the Spatial Transformer Network (STN)~\cite{jaderberg2015spatial}. For every vector in step \textbf{1)}, we construct its $3\times3$ neighbor vectors according to the transformer grid defined in step \textbf{2)}. For the points with only 7 adjacent points, the missing vector is set to be $\textbf{0}$. See the third row of Fig.~\ref{fig:module} for illustration.

\textbf{4) Run the standard $3\times3$ convolution.} Since the transformed vectors have been obtained in the above step, we can use standard $3\times3$, stride $1\times3$ convolution kernels to process it directly. For example, for the feature vectors $U\in\mathbb{R}^{1\times 768\times C1}$ in level-\emph{3}, it is transformed to the vectors $V\in\mathbb{R}^{1\times 768\times C2}$ after the processing of $conv3\times3(C1,C2)$ with stride $1\times3$.

\begin{figure*}[tb!]
  \centering
  \includegraphics[width=\textwidth]{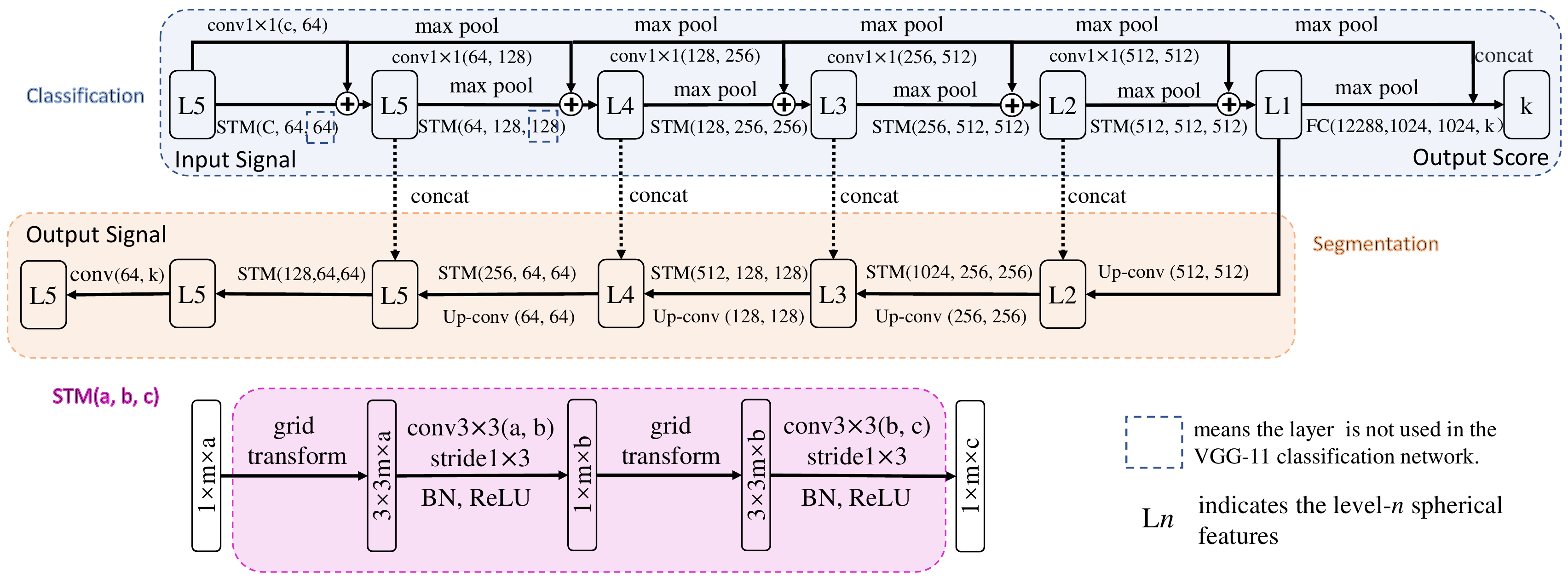}
  \vspace{-4mm}
  \caption{Overview of the model architecture for classification and segmentation. Our spherical transformer module is implemented according to Fig.~\ref{fig:module}. Up-conv uses standard $2\times2$ deconvolution function. For classification, we change the basic VGG-11 model and add $1\times1$ convolution kernel to improve the anti-rotation ability. For fully connected layer in classification compared to it in standard VGG-11 architecture, we use 1024 dimensions rather than 4096. And for segmentation, we use the U-Net architecture without the $1\times1$ convolution.}
   \vspace{-6mm}
  \label{fig:architecture}
\end{figure*}

\vspace{-3mm}
\paragraph{Spherical pooling layer.}
In HEALPix spherical grid, the distribution of spherical points for each level resolution is fixed. For example, in Fig.~\ref{fig:module}, the level-2 spherical points are marked as plus sign. Its four nearest points in level-3 grid are marked as dots. In the yellow area in HEALPix gird, these four points are adjacent to each other. See the last row of Fig.~\ref{fig:module}. The nested points index has defined a natural pooling way that four successive vectors of level-\emph{3} are pooled to form a vector $p$ of level-\emph{2}. Thus we can use $1\times4$ max pooling layer directly instead of building spherical transformer grid for spherical pooling layer. Consequently, the spherical features in level-\emph{l} are straightforward pooled to the features in level-\emph{l-1} as they are coded in proper order in HEALPix grid.

\vspace{-3mm}
\subsection{Network architecture}
\vspace{-1mm}
As described in Fig.~\ref{fig:module}, we have constructed our spherical convolution layer and spherical pooling layer by using the proposed spherical transformer method. By doing this, we can directly use the classical CNN architectures as we originally designed. For 3D object classification, we adopt VGG-11 network architecture but replace its convolution layer and pooling layer with our proposed spherical convolution and pooling layer. And to get rotational-equivariant, we change the basic VGG-11 model and more details are described in next section. For semantic segmen\-tation, we similarly construct a novel network which combines the U-Net architecture with the proposed STM. Detailed schematic for these two architectures is shown in Fig.~\ref{fig:architecture}. Moreover, both the classification and segmentation networks share a common encoder architecture. Since these architectures have been verified in various vision tasks, it is reasonable to apply them to the proposed spherical structured grids.
\begin{table}[tb!]
  \centering
  \caption{Classification accuracy on the spherical MNIST dataset for validating STM.\centering}
  \label{acc_table1}
  \begin{tabular}{lcr}
    \toprule
    Method     & Acc ($\%$)     &  Parameter \# \\
    \midrule
    S2CNN~\cite{s.2018spherical}           & 96.00   & 58k  \\
    UGSCNN~\cite{jiang2018spherical}            & 99.23   & 62k  \\
    \midrule
    STM (Ours)             & 99.36   & 32k     \\
    \bottomrule
  \end{tabular}
    \vspace{-5mm}
\end{table}

\vspace{-3mm}
\paragraph{Anti-rotation model}
As previous studies e.g. S2CNN~\cite{s.2018spherical} and SphericalCNN~\cite{10.1007/978-3-030-01261-8_4} propose spherical convolutions that are rotational-equivariant, in order to improve the anti-rotation ability of our 3D object classification model, we modify our VGG-11 liked model architecture as shown in Fig.~\ref{fig:architecture}. It is well known that convolutional neural networks implement translation-equivariant by construction, for other transformations such as rotations, however, they are ineffective. Though recently some convolution kernels are proposed for rotation-equivariant such as G-CNNs~\cite{cohen2016group} and SFCNNs~\cite{weiler2018learning}, here we resort to the simple $1\times1$ convolution kernel which is naturally rotation-equivariant.
For its simplicity, we add $1\times1$ convolution kernel to our basic model. To find features that are sensitive to these two convolutions, for every $3\times3$ convolution layer before max pooling, we sum these two features. Certainly, to preserve most rotation-equivariant features, we concatenate these two features before the first full-connected layer. More details can be seen in Fig.~\ref{fig:architecture}.
\vspace{-3mm}
\section{Experiments}
\label{sec:experiment}
\vspace{-2mm}
\subsection{Spherical MNIST}
To verify its efficiency, we first use our method on classic digital recognition.

\vspace{-4mm}
\paragraph{Experiment setup.}
To project digits onto the surface of the sphere, we follow the projection method of S2CNN~\cite{s.2018spherical} and UGSCNN~\cite{jiang2018spherical}. We benchmark our method with the above two spherical CNNs. All models are trained and tested on data that has not been rotated. In this experiment, we use a 5 layers CNN architecture with 4 conv-pool-BN-ReLU and 1 FC-softmax. 

\vspace{-4mm}
\paragraph{Results and discussion.}
Table~\ref{acc_table1} shows the classification accuracy on spherical MNIST. It shows that our method outperforms S2CNN and UGSCNN. In particular, the number of parameters of our model is about only half of that of the above models. We attribute our success to the seamless reuse of 2D CNNs which has been dominant in wide range of vision tasks.
\vspace{-4mm}
\subsection{3D object classification}
\begin{figure*}[tb!]
  \centering
  \includegraphics[width=\textwidth]{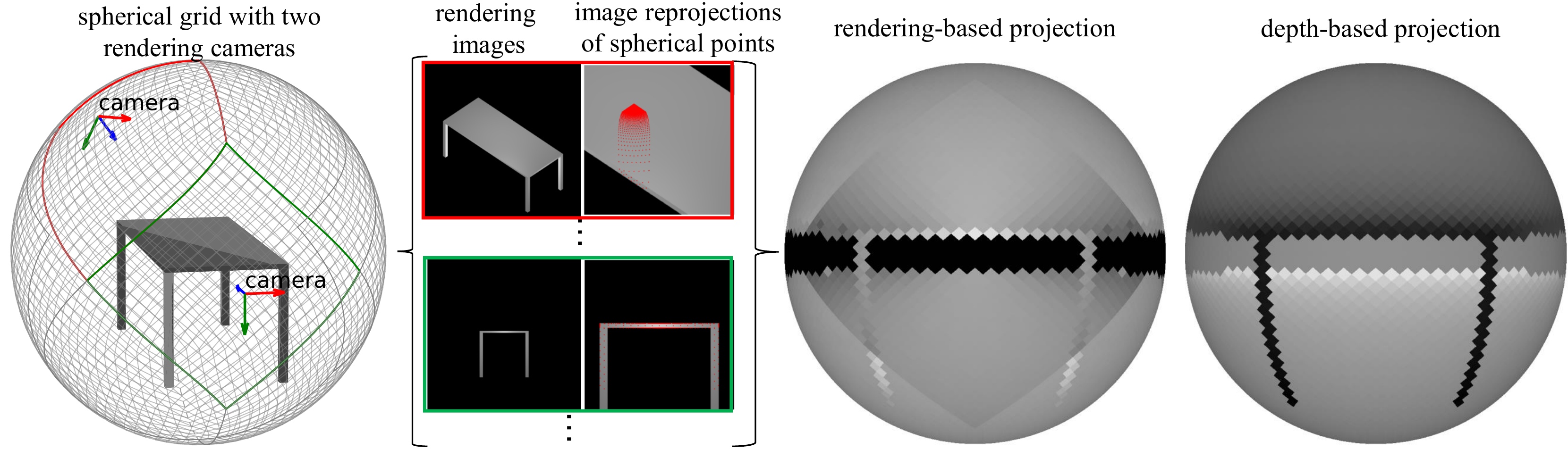}
  \caption{Two projection methods used in our experiments, including the proposed rendering-based one. This special rendering-based projection allows to directly combine the two methods' features.}
  \vspace{-2mm}
  \label{fig:projection}
\end{figure*}

The benchmark dataset used in this task is ModelNet40~\cite{Wu20153DSA}. It contains 12,311 shapes across 40 categories, which is used to illustrate the applicability of our approach in 3D deep learning. For this study, we focus on classification accuracy and rotational-equivariant. Two types of spherical projection in our experiments are used: depth-based projection and rendering-based projection. 
\vspace{-4mm}
\paragraph{Depth-based projection.}
For depth-based projection, we follow the processing protocol of \cite{s.2018spherical}. Specifically, as shown in Fig.~\ref{fig:projection}, we first move the model to the origin and then normalize it. Then we calculate level-\emph{5} resolution spherical points and send a ray towards the origin for each point. Three kinds of information from the intersection are obtained: ray length, $sin$ and $cos$ values of the angle between the surface normal of the intersected object point and the ray. The data is further augmented by using the convex hull of the input model, finally forming spherical signals with 6 input channels. The rightmost plot in Fig.~\ref{fig:projection} shows the visualized depth-based spherical signals. Even if spherical image is one kind of feature representation from 3D to 2D, we can clearly see the various parts of the table, just like a distorted image. 

\vspace{-4mm}
\paragraph{Rendering-based projection.}
Motivated by the multi-view based method~\cite{Kanezaki2018RotationNetJO} that has achieved the best performance on ModelNet40 dataset, we explore and propose a render\-ing-based spherical projection method. However, such rendering-based methods have difficulty in stitching multi-view images to a spherical image directly. This is because the projections of multi-view images on spherical image are not aligned. One 3D object point can even appear multiple times in stitched spherical image. Therefore, we propose a novel projection strategy to separately obtain projections of 12 regions whose inner points correspond to different parts of the object, while boundary points correspond to the same part. For example, in Fig.~\ref{fig:projection}, the red and green regions are 2 of 12 regions divided by HEALPix grid, respectively. We put the virtual camera on the ray of the origin and the center of this region which captures the image of the model from the current perspective. By adding six fixed point light sources on $+-x$, $+-y$, $+-z$ axes, 12 gray images can be rendered. Through the depth-based projections, we already know the point at which the model intersects the spherical ray. Hence, the gray value of one spherical point can be obtained by re-projecting the corresponding object point back to the rendered image of each region. It can be found that our rendering-based method only uses the internal points of the object, which contains the local shading of the object surface, without contour information. The third sub-figure from left to right in Fig.~\ref{fig:projection} shows the rendering-based spherical images from the view of the camera of the green region. Although the rendering spherical image is not visually straightforward, it indeed provides alternative feature that improves the performance.

\begin{table*}[tb!]
  \centering
  \caption{Classification accuracy (Acc ($\%$) and SO(3)/SO(3) Acc ($\%$)) on ModelNet40.\centering}
  \label{acc_table}
	\resizebox{0.7\textwidth}{!}{
  \begin{tabular}{llcc}
    \toprule
    Method     & Input     & Acc & SO(3)/SO(3)\\
    \midrule
    MVCNN 12x \cite{su15mvcnn} & images      & 89.5    & 77.6  \\
    3D Shapenets \cite{Wu20153DSA} & voxels      & 84.7   & -   \\
    Voxnet \cite{maturana2015voxnet}       & voxels      & 83.0  & 87.3     \\
    PointNet~\cite{qi2017pointnet}           & points      & 89.2   & 83.6\\
    PointNet++ \cite{qi2017pointnet++}         & points      & 91.9  & 85.0 \\
    DGCNN \cite{wang2018dynamic}              & points      & 92.2   & - \\
    SFCNN \cite{rao2019spherical}         & points      & 92.3  & 91.0 \\
    S2CNN \cite{s.2018spherical}           & spherical   & 85.0  & - \\
    SphericalCNN~\cite{10.1007/978-3-030-01261-8_4}       & spherical   & 88.9   & 86.9\\
    UGSCNN~\cite{jiang2018spherical}            & spherical   & 90.5  & - \\
    \midrule
    Ours-VGG11 (depth)               & spherical   & 92.3    & 87.7  \\
    Ours-rot (depth)               & spherical   & 92.7    & 91.3  \\
    Ours-VGG11 (rendering, w/o pre-training)            & spherical   & 86.2 & -   \\
    Ours-VGG11 (rendering, w/ pre-training)            & spherical   & 91.1  & -  \\
    Ours (overall)             & spherical   & 93.0    & - \\
    \bottomrule
  \end{tabular}}
  \vspace{-3mm}
\end{table*}

\vspace{-3mm}
\paragraph{Experiment setup.}
For this task, the level-\emph{5} spherical resolution is used. We use two spherical inputs to train the network separately. For classification accuracy, we use the aligned ModelNet40 data~\cite{sedaghat2016orientation}. While SO(3)/SO(3) means trained and tested with arbitrary rotations. Our rendering-based projection method can use the VGG-11 model parameters pre-trained on the ImageNet data. For classification accuracy, we compare the best result of our method with other 3D deep learning methods. The baseline algorithms we choose include UGSCNN~\cite{jiang2018spherical}, PointNet++~\cite{qi2017pointnet++}, VoxNet~\cite{maturana2015voxnet}, S2CNN~\cite{s.2018spherical}, SphericalCNN~\cite{10.1007/978-3-030-01261-8_4}, SFCNN \cite{rao2019spherical} and MVCNN~\cite{su15mvcnn}. 

\vspace{-4mm}
\paragraph{Results and discussion.}
Table~\ref{acc_table} shows the classification accuracy on ModelNet40. The proposed spherical depth feature based model is superior to the above existing methods. And our rotational-equivariant model shows better performance in both the best classification accuracy and SO(3)/SO(3) classification accuracy. Our rendering-based approach also achieves promising results after using pre-trained parameters. The overall model means we combine the features of the penultimate full-connection layer of the two methods and that we retrain the last fully connection layer. When combining depth-based projection and rendering-based projection, our overall method achieves the highest accuracy. This also suggests that our devised rendering-based projection is effective and complementary to depth-based projection method.
\begin{figure*}[tb!]
  \centering
  \includegraphics[width=\textwidth]{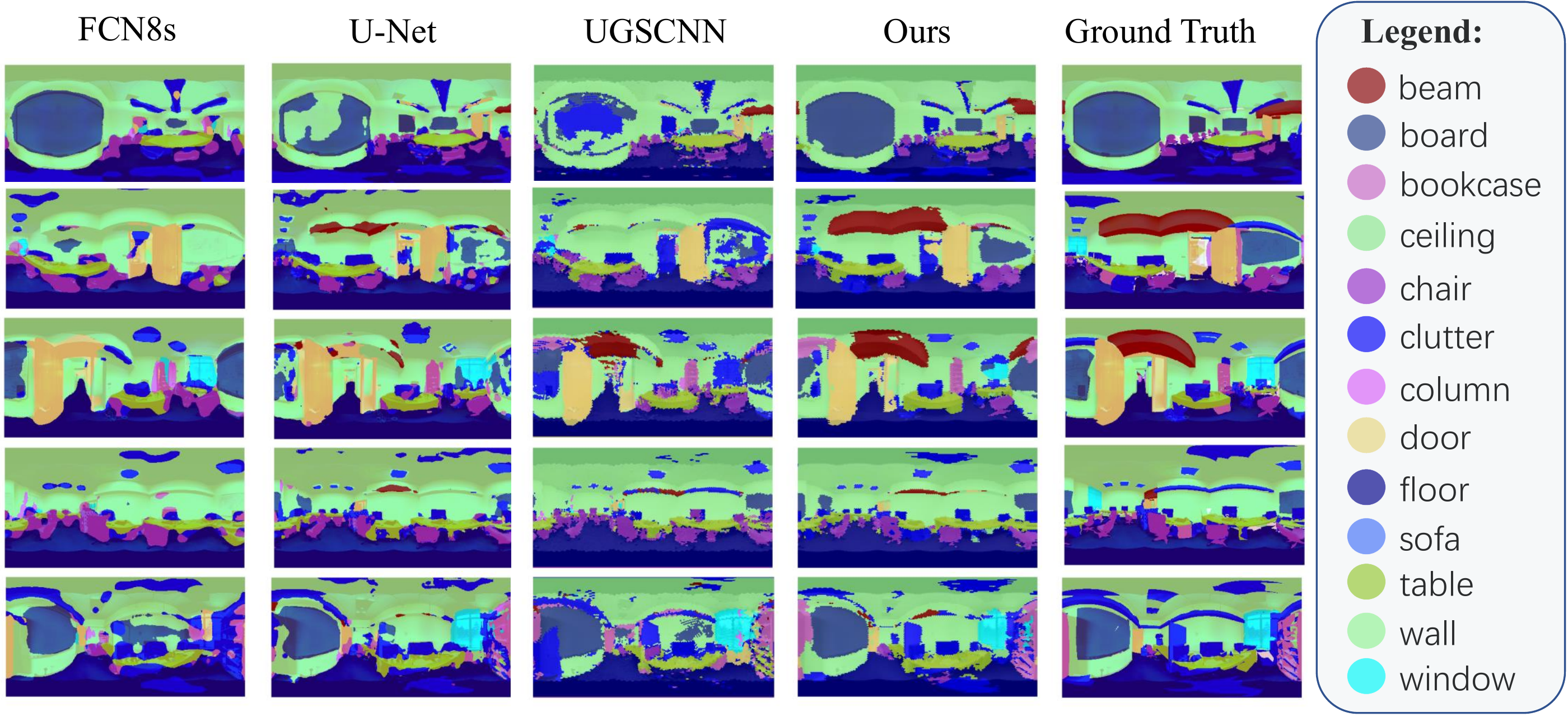}
  \vspace{-6mm}
  \caption{Semantic segmentation results on Stanford 2D3DS test dataset. Our results are generated on a level-\emph{5} HEALPix grid and mapped to the planar image by using nearest neighbor sampling for visualization. Each row shows the results of different methods including ground truth for a specific scene. Note that only our method well captures the beam structure.}
  \label{fig:segresult}
\end{figure*}
\vspace{-3mm}
\subsection{Spherical image semantic segmentation}
We demonstrate the semantic segmentation capability of the proposed spherical transformer module on the spherical image semantic segmentation task. We use the Stanford 2D3DS dataset~\cite{2017arXiv170201105A} for comparison with the state-of-the-art UGSCNN~\cite{jiang2018spherical}. The 2D3DS dataset contains a total of 1,413 equirectangular RGB images, along with their corresponding depths, and semantic annotations across 13 different classes. Except compared with UGSCNN that is aimed to segment spherical image, we also include classic 2D image semantic segmentation networks for more comprehensive comparison. 

\begin{figure}[tb!]
  \centering
  \includegraphics[width=0.45\textwidth]{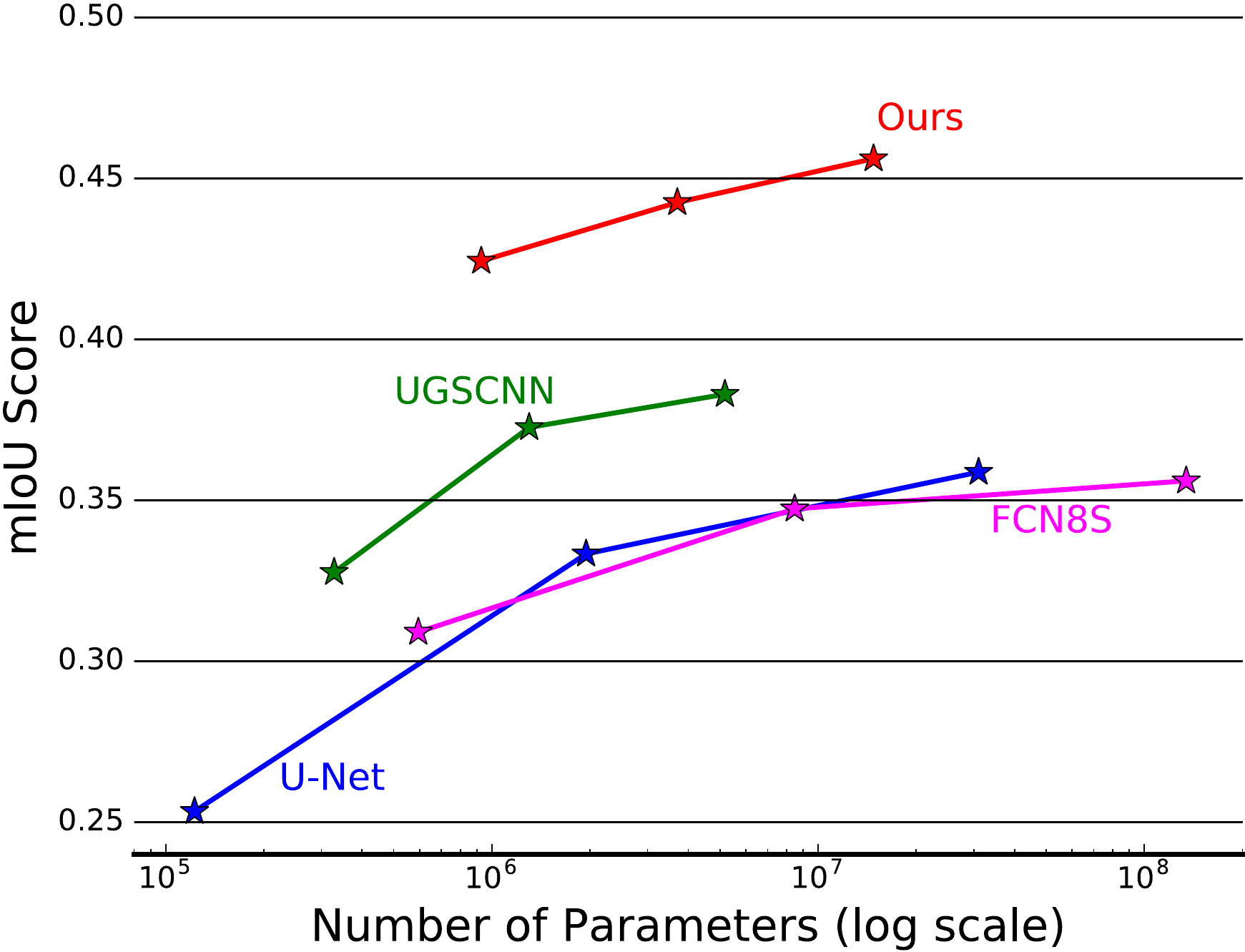}
  \vspace{-2mm}
  \caption{Parameter efficiency study on 2D3DS for semantic segmentation. Our proposed spherical segmentation model outperforms UGSCNN and two planar-based methods by notable margin across all parameter regimes.}
  \vspace{-6.8mm}
  \label{fig:seg}
\end{figure}
\vspace{-4mm}
\paragraph{Experiment setup.} 
To apply our method on semantic segmentation tasks, the first thing is sampling on the original rectangular images to obtain spherical signals. To make a fair comparison with UGSCNN~\cite{jiang2018spherical}, we follow the interpolation method of UGSCNN. The input RGB-D channels are interpolated using bilinear interpolation, while semantic segmentation labels are acquired using nearest-neighbor interpolation. We use level-\emph{5} resolution and the official 3-fold cross validation to train and evaluate the experimental results. For this task, we benchmark our semantic segmentation results against the spherical semantic segmentation architecture UGSCNN and two classic semantic segmentation networks: U-Net~\cite{ronneberger2015u} and FCN8s~\cite{long2015fully}. We evaluate the performance under two standard metrics: mean Intersection-over-Union (mIoU), and pixel-accuracy. These methods are compared under two settings: peak performance and parameter efficiency study by varying model parameters. For parameter efficiency study, we change the number of parameters by reducing the dimension of the convolution layer. The only difference between our network and the standard U-Net is the dimension of the last layer of our model encoder is 512, rather than 1024, as the ladder case does not have obvious improvement. 
\vspace{-2mm}
\paragraph{Results and discussion.}
Figure~\ref{fig:seg} compares our model performance against state-of-the-art baselines. Our proposed spherical segmentation network significantly outperforms UGSCNN and two planar baselines over the whole parameter range. Here three different parameter numbers for each method denote the reduction of feature dimensionality, in line with the setting in UGSCNN. Fig.~\ref{fig:segresult} shows a visualization of our semantic segmentation results compared to the ground truth, UGSCNN and two planar baselines. It can be seen that our method clearly achieves the best accuracy and the results are also visually appealing.
\vspace{-3mm}
\section{Conclusion}
\vspace{-1mm}
\label{sec:conclusion}
In this paper, we present a novel method to transform the spherical signals to structured vectors that can be processed through standard CNNs directly. 
Experimental results show significant improvements upon a variety of strong baselines in both tasks for 3D object classification and spherical image semantic segmentation. 
With the increasing availability and popularity of omnidirectional sensors 
we believe that the demand for specialized models for spherical signals will increase in the near future. 
\vspace{-2mm}

%
%
%
{\small
\bibliographystyle{splncs04}
\bibliography{mybibliography}
}

\end{document}